# Reconstruction and Discrete-Event Simulation Assessment of an Analytical Cycle-Time Model for Intra-Aisle Robotic Picking

**Mohsen Reda Mahmoud Moursi**[1], **Ahmed Stohy**[2,*]

[1] Graz University of Technology, Graz, Austria

[2] TU Wien, Vienna, Austria

*Research conducted while Mohsen Reda Mahmoud Moursi was a Master's student.*

**Email:** mohsenredamahmoud@gmail.com; ahmed.stohy@student.tuwien.ac.at

***Corresponding author:** Ahmed Stohy, ahmed.stohy@student.tuwien.ac.at

## Abstract

This article presents an analytical–DES benchmark for intra-aisle robotic picking in automated warehousing. The investigated cycle combines horizontal shuttle travel, tote handling and item picking at storage locations. Simulation results are compared with the analytical formulation of Lerher et al. under the same rack geometry, shuttle-motion, handling and picking conditions and the same definitions of cycle time and derived item throughput. Reproducing the published analytical tables requires the distance interpretation $d_{ES}$ = 1.5 m and $d_{TB}$ = 0 m. The analytical–DES benchmark comprises 180 scenarios formed from three rack lengths, two paired motion profiles and n = 1–30 visited picking locations, with ten simulation replications per scenario. The six n = 1 cases are reported separately because no travel-between component is present. Across the remaining 174 scenarios, the mean absolute relative cycle-time discrepancy is 0.8655%, the median is 0.6758%, and the maximum is 7.0370%. The mean absolute error is 2.411 s and the RMSE is 2.744 s. The simulated scenario-mean cycle time is below the analytical value in all 174 principal scenarios. The results therefore show close numerical correspondence under the tested conditions, but not exact agreement, formal statistical equivalence, or validation against an operating warehouse.

## 1. Introduction

Warehousing connects inbound supply, storage, order fulfilment and outbound distribution. Order picking converts customer or production demand into physical item retrieval and affects response time, capacity, accuracy and resource use. E-commerce can involve small orders, broad product assortments, short response times and variable demand, which increases the importance of storage, picking, routing and control decisions (de Koster et al., 2007; Boysen et al., 2019). Automated order-picking systems differ in where items are picked, which resource moves them and how storage and transport are organised. In the intra-aisle concept studied here, a mobile resource travels within the aisle and an integrated picking function transfers items at the storage location instead of sending all goods to a separate workstation.

Analytical and simulation models support different parts of warehouse performance evaluation. Analytical models use equations to relate geometry, motion and handling assumptions to cycle time or throughput. Discrete-event simulation represents state changes at events such as travel, tote transfer, picking or waiting and can include control rules, resource interactions and random variation. Recent warehouse simulation studies continue to use DES for automated storage and shuttle systems (Saini et al., 2025; Ferrari et al., 2025), while analytical-to-simulation comparisons have been reported for shuttle-based and robotic compact storage/retrieval systems (Eder, 2024; Trost and Eder, 2024). Close agreement between two models can support verification and interpretation, but it does not by itself show that either model represents an operating warehouse (Sargent, 2013; Ferrari et al., 2024).

Lerher et al. (2025) provide equations for the expected cycle time and item throughput of a multi-command picking cycle performed by an autonomous mobile robot with an integrated picking robot. The analysed cycle is limited to one aisle and one storage level. The mobile resource starts at the front of the aisle, visits n picking locations, performs tote handling and item picking, and returns to the front. The published study uses rack lengths of 30, 60 and 90 m, two paired velocity/acceleration profiles and n = 1–30 visited picking locations. This provides the analytical reference used for the single-level comparison in the present article.

For a direct comparison, the analytical calculation and simulation must use the same rack geometry, motion parameters, handling and picking times and the same cycle-time and throughput boundaries. The simulation still represents individual location-specific movements rather than replacing its travel logic with the analytical equations. A separate reproducibility point arises in the published distance notation. Lerher et al. use the same $d_x$ notation for the one-way and travel-between components without separate numerical values. Checking the equations against the published result tables gives $d_{ES}$ = 1.5 m and $d_{TB}$ = 0 m. Those values are therefore stated explicitly before the analytical–simulation comparison.

The research question addressed in this article is: How closely do the cycle-time and derived item-throughput results from the simulation model correspond to the analytical reference when rack geometry, motion and handling parameters and cycle-time and throughput definitions are aligned, and how does the discrepancy change with rack length, motion profile and the number of visited picking locations?

The article has four objectives. First, it reconstructs the single-level analytical reference and states the distance interpretation needed to reproduce the published result tables. Second, it documents how the simulation cycle is aligned with the analytical cycle while retaining location-specific movement and controller-determined sequencing. Third, it evaluates all 180 benchmark scenarios using ten replications per scenario and reports replication-based uncertainty. Fourth, it quantifies both the size and direction of the remaining analytical–DES discrepancy. The scientific contribution is therefore not simply an implementation check: it is a reproducible benchmark of model-form correspondence between an expected-distance analytical representation and an explicitly executed DES representation, including the systematic direction of the residual difference, its dependence on operating parameters, and the structurally distinct n = 1 boundary case. The comparison is intended to assess model correspondence rather than establish statistical equivalence or physical-system validity.

## 2. Background and Analytical Reference

### *2.1 Analytical and simulation-based performance evaluation*

Analytical performance models describe selected relationships between a warehouse architecture, its assumptions and measures such as travel time, cycle time, waiting and throughput. The equations and processes that matter depend on the architecture. In a tier-captive shuttle-based storage and retrieval system, shuttle travel, lift service, buffers and storage allocation affect performance (Eder, 2020; Tappia et al., 2019). Robotic compact storage/retrieval systems require different representations for grid travel, stack access, relocation and picking-station capacity (Trost and Eder, 2024). This architecture dependence is why the process boundary has to be stated before analytical and simulation results are compared.

In a discrete-event simulation, the model state changes when events occur. Simulation is useful when activities interact and when timing or choices can vary between runs. More model detail does not by itself show that the model is correct for its intended use. Verification checks whether the planned model has been implemented correctly, while validation concerns whether the model is adequate for the purpose for which it will be used (Sargent, 2013). Comparison with an analytical calculation can reveal implementation or definition differences, but agreement with another model is not the same as agreement with measurements from a real warehouse.

Clear reporting is therefore part of the assessment. The STRESS guidelines emphasise reporting the study objectives, inputs, conceptual model, implementation, experimentation and results so that a simulation study can be understood and, where possible, reproduced (Monks et al., 2019). The analytical formulation, simulation boundary, experiment design, aggregation rules and statistical analysis are reported separately here for the same reason.

### *2.2 Analytical reference for intra-aisle robotic picking*

The analytical reference represents an AMR with an integrated picking robot performing a multi-command cycle in a two-sided storage-rack aisle (Lerher et al., 2025). The physical concept may be used in a multi-level rack, but the analysed cycle is limited to one aisle and one storage level. Its assumptions include random and equally likely location selection, a starting and return point at the front of the aisle, known horizontal velocity and acceleration, and fixed tote-handling and item-picking times. The calculation includes horizontal travel, tote handling and item picking. Replenishment, waiting for an order tote, vertical tote movement and competition for shared external resources are outside this analytical cycle.

The inputs used in the reconstruction are rack length $L_{SR}$, horizontal velocity $v_x$, acceleration $a_x$, the number of visited picking locations n, tote pick-up/set-down time $t_{P/S}$ and item-picking time $t_{OP}$. The values used by Lerher et al. (2025) and reproduced here are summarised in Table 1.

Table 1. Analytical reference parameters.

| Parameter | Symbol | Values used |
|---|---|---|
| Rack length | $L_{SR}$ | 30, 60, 90 m |
| Motion profile M1 | ($v_x$, $a_x$) | (2.5 m/s, 1.5 m/s²) |
| Motion profile M2 | ($v_x$, $a_x$) | (4.0 m/s, 3.0 m/s²) |
| Visited picking locations | n | 1–30 |
| Tote pick-up/set-down time | $t_{P/S}$ | 3.4 s per occurrence |
| Item-picking time | $t_{OP}$ | 10 s per visit |
| Additional one-way distance | $d_{ES}$ | 1.5 m in the reconstruction |
| Additional travel-between distance | $d_{TB}$ | 0 m in the reconstruction |

The value n is the total number of visited picking locations in one multi-command cycle. The auxiliary value j = (n/2) is used in the distance equations and is not rounded for odd n. Because the published formulation uses the same $s_x$ and $d_x$ symbols for both horizontal travel cases, the reconstruction writes separate one-way and travel-between terms, $s_{ES}$ and $s_{TB}$:

$$s_{ES} = \frac{L_{SR}}{\left(\frac{n}{2}\right)+1} + d_{ES} \quad \textbf{(1)}$$

$$s_{TB} = \frac{L_{SR}}{\left(\frac{n}{2}\right)+1} + d_{TB} \quad \textbf{(2)}$$

Horizontal motion follows the same triangular/trapezoidal kinematic relation for both travel components. For a movement distance s, maximum velocity $v_x$ and acceleration $a_x$, the motion time is

$$t_x(s; v_x, a_x) = \begin{cases} \frac{s}{v_x} + \frac{v_x}{a_x}, & s > \frac{v_x^2}{a_x}, \\ 2\sqrt{\frac{s}{a_x}}, & s \le \frac{v_x^2}{a_x}. \end{cases} \quad \textbf{(3)}$$

The one-way and travel-between travel times are E(ES) = $t_x(s_{ES}; v_x, a_x)$ and E(TB) = $t_x(s_{TB}; v_x, a_x)$, respectively. In the reconstruction used here, the motion-time function is applied to the expected distance, t(E[S]), rather than averaging motion times over realised distances, E[t(S)]. This choice is retained because it reproduces the published tables.

The analytical multi-command cycle time is

$$CT_A(n) = 2E(ES) + (n - 1)E(TB) + n(2t_{P/S} + t_{OP}) \quad \textbf{(4)}$$

Each analytical visit contains one tote pick-up, one item-picking operation and one tote set-down. With $t_{P/S}$ = 3.4 s and $t_{OP}$ = 10 s, the handling and picking contribution is 16.8 s per visited location. The corresponding item-throughput measure is calculated from the unrounded analytical cycle time as

$$TH_A(n) = \frac{3600\,n}{CT_A(n)} \quad \textbf{(5)}$$

$TH_A$ is reported in items per hour and is a value derived from the analytical cycle definition. It is not a directly counted simulation output.

### *2.3 Reconstruction of the published analytical results*

Lerher et al. (2025) use the same $d_x$ notation for the one-way and travel-between distance relations and do not give separate numerical values. Using one common value for both terms does not reproduce all reported results. Checking the equations against Tables II–VII gives $d_{ES}$ = 1.5 m and $d_{TB}$ = 0 m. The n = 1 cases determine the one-way term because the multiplier of E(TB) is zero; the n > 1 cases then determine the travel-between term. The value $d_{TB} = 0$ removes only the additional distance offset. The travel-between distance and travel time remain positive for n > 1.

These values reproduce the published cycle-time values and the published throughput values after rounding. They are used as the analytical reference for the benchmark. The reconstruction is reported as the numerical interpretation required for reproducibility; it is not presented as a correction of the original publication.

## 3. Discrete-Event Simulation Model and Analytical–Simulation Alignment

### *3.1 DES representation*

The simulation model was developed in Siemens Tecnomatix Plant Simulation version 24.4.9.3160 using SimTalk control methods. The complete model can represent multiple aisles, two storage levels, tote routing and replenishment, but this article uses only the single-level analytical–DES benchmark. The benchmark uses the geometry required for the analytical-reference comparison. Replenishment, vertical tote movement, inbound and outbound lift service, external conveyor movement and other processes outside the measured shuttle cycle are not part of the benchmark cycle result.

The benchmark uses a predefined storage state. For each order, the requested number of eligible positive-stock storage locations is selected randomly without selecting the same location twice. The selected tasks are linked to one order tote and released for shuttle execution. The random step therefore concerns the set of selected locations. During benchmark execution, the implemented side/column ranking rule determines the next eligible task. Thus, storage locations are selected randomly without replacement, whereas their executable visit sequence is determined by the controller rule.

### *3.2 Horizontal travel, handling and picking*

For a selected task, the controller calculates the target horizontal position from the storage column and the travel distance from the current shuttle position. The configured velocity and acceleration are then used with the same triangular/trapezoidal motion calculation as in the analytical formulation. At the storage location, the model applies tote-handling and item-picking times and records the corresponding travel, handling and picking components. After the required picking stops, the return movement uses the same horizontal motion calculation.

The analytical and simulated travel representations remain different in one important respect. The analytical model converts expected one-way and travel-between distances into motion times. The simulation executes the location-specific movements produced by the selected locations and the controller sequence. The common kinematic parameters are aligned, but the individual distances are not replaced by their analytical expected values.

### *3.3 Cycle boundary and KPI construction*

For the benchmark, the measured simulated cycle starts when the shuttle leaves its home position and ends when it returns after the required picking stops. It includes shuttle travel, tote handling and item picking. The inbound and outbound lifts, external conveyors, waiting outside the shuttle trip and replenishment are not included. The complete cycle time is the sum of the recorded time components rather than the difference between two external timestamps.

Each replication is a finite-horizon benchmark run that starts from the predefined fully occupied single-level rack state; no warm-up period is removed because the measured quantity is a complete shuttle-cycle mean rather than a steady-state capacity estimate. The benchmark task generator is configured with a maximum of 1,000 orders per replication, and the required n eligible positive-stock locations for each order are sampled without replacement. Replications use separate random-number streams. At the end of a replication, only complete cycles are accepted: the recorded cycle must contain n pick rows, a picked quantity of n, one completed drop-off record, a positive cycle time, and n distinct storage-location keys. A cycle that has pick activity but no completed drop-off when the run ends is classified as an end-cutoff cycle and excluded. The accepted complete cycles are averaged to obtain one replication mean; the ten replication means are then averaged to obtain $CT_S$. Individual cycles from the same replication are not treated as separate independent observations. The sample standard deviation and standard error are calculated from the ten replication means, and a two-sided 95% Student-t confidence interval is calculated with nine degrees of freedom (Law, 2020).

Simulation-derived item throughput is calculated once from the scenario-mean cycle time as

$$TH_S(n) = \frac{3600\,n}{CT_S(n)} \tag{6}$$

$TH_S$ is calculated once from the scenario-mean cycle time. It is not the mean of throughput values calculated for individual cycles or replications. Both $TH_A$ and $TH_S$ are therefore derived from their corresponding cycle times and should not be interpreted as independent throughput observations.

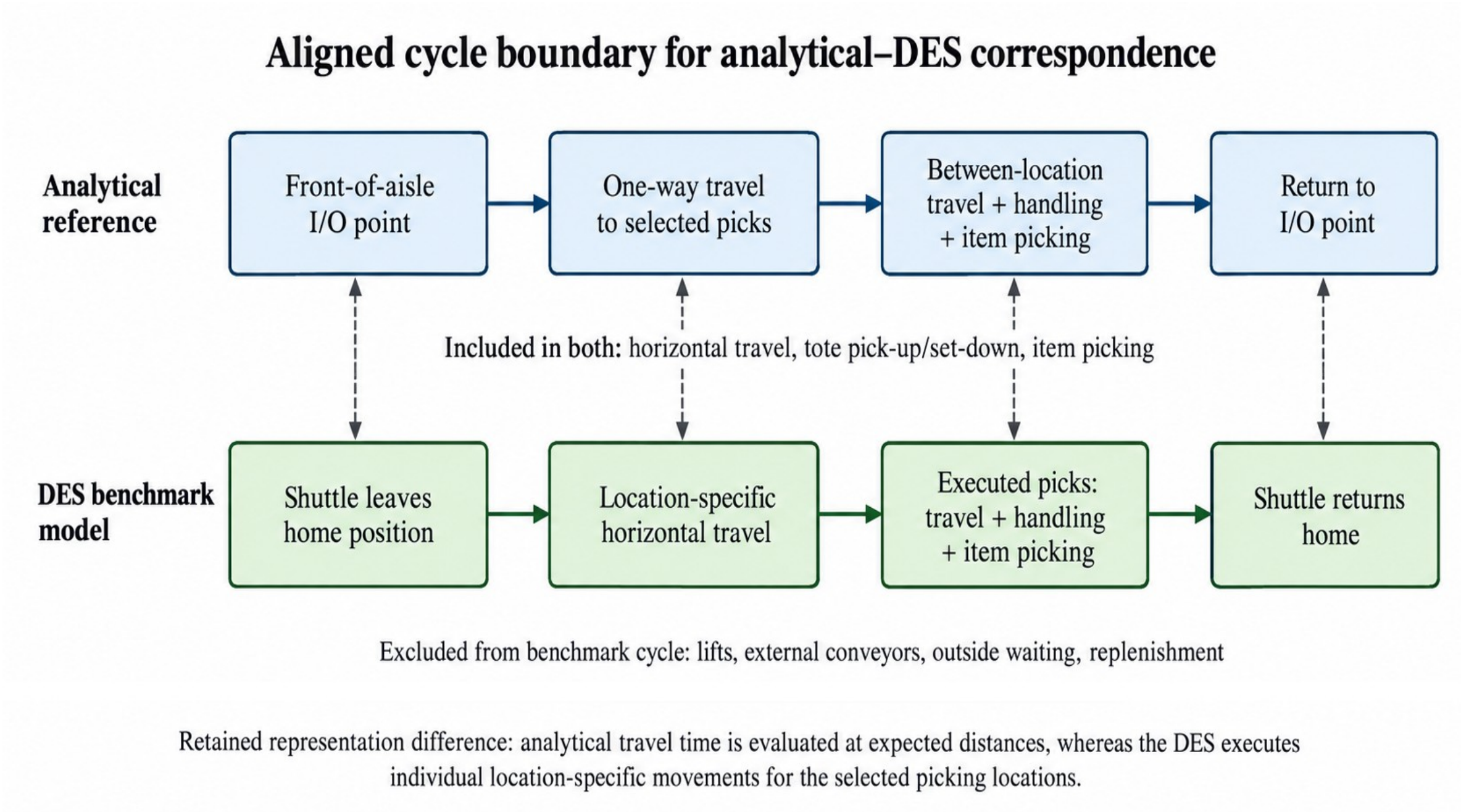

*Figure 1. Aligned cycle boundary for analytical–DES correspondence. The same process components are included in both representations, while lifts, external conveyors, outside waiting and replenishment are excluded. The analytical reference evaluates travel at expected distances, whereas the DES executes location-specific horizontal movements for the selected picking locations.*

### *3.4 Analytical–DES alignment*

Table 2 summarises how the rack geometry, paired motion parameters, handling and picking times, cycle boundary and KPI definitions are aligned. The purpose of these controls is to compare the two representations under the same stated conditions without removing the simulation's individual location-specific movements.

*Table 2. Analytical–DES alignment for the analytical–DES benchmark.*

| Element | Analytical reference | DES benchmark | Alignment status |
|---|---|---|---|
| System scope | One aisle, one storage level | Single-level comparison configuration | Aligned |
| Rack length | 30, 60, 90 m | 30, 60, 90 m | Aligned |
| Motion profile | M1 or M2 | Same paired v/a settings | Aligned |
| Visited locations | n = 1–30 | n = 1–30 | Aligned |
| Location selection | Random and equally likely assumption | Eligible locations sampled randomly without replacement | Closely aligned |
| Visit sequencing | Represented through expected-distance formulation | Implemented side/column ranking | Representation differs |
| Horizontal motion | Triangular/trapezoidal relation | Same relation for realised distances | Aligned kinematics |
| Tote handling | 3.4 s pick-up + 3.4 s set-down per visit | Same values | Aligned |
| Item picking | 10 s per visit | 10 s per visit | Aligned |
| Cycle boundary | Front of aisle to return to front | Leave home to return home | Aligned process boundary |
| External lifts/conveyors | Excluded | Excluded from measured benchmark cycle | Aligned |
| Replenishment | Excluded | Not generated in benchmark | Aligned |

| Element | Analytical reference | DES benchmark | Alignment status |
|---|---|---|---|
| Travel representation | Motion time at expected distances | Individual location-specific movements | Intentionally different |

### *3.5 Verification*

Verification checks that the benchmark uses the intended settings and that the stored results match the experiment design. The final files contain 180 unique scenarios and 1,800 unique scenario–replication records, with ten replications for every scenario. The clean dataset contains no duplicate scenario–replication keys and no invalid cycles. Recalculating the scenario means, sample standard deviations, standard errors and 95% confidence intervals from the replication file reproduces the stored scenario results to numerical precision. These are implementation and data-consistency checks; they are not a validation against an operating warehouse.

## 4. Experimental Design and Statistical Analysis

### *4.1 Experimental design*

The analytical–DES benchmark contains 180 scenarios arranged in six rack-length and motion-profile combinations over n = 1–30 visited picking locations. Rack length takes the values 30, 60 and 90 m. The two motion profiles are 2.5 m/s with 1.5 m/s² and 4.0 m/s with 3.0 m/s². Velocity and acceleration are used as paired profiles rather than as separately crossed factors. Each scenario has ten simulation replications, giving 1,800 replication-level results. The benchmark uses the same fully occupied initial rack state within the comparison, no replenishment, and the finite-horizon complete-cycle rule described in Section 3.3.

The six n = 1 scenarios are reported separately because no travel between picking locations occurs when only one location is visited. The principal correspondence analysis therefore contains the 174 scenarios with n = 2–30, and the replication-level factor analysis uses the corresponding 1,740 replication results. Across all 1,800 replications, the final clean dataset contains 5–133 accepted complete cycles per replication (median 11; interquartile range 7–19.25). No invalid cycles were retained; 1,622 replications contain one incomplete end-cutoff cycle that is excluded from the replication mean.

Table 3. Experimental design.

| **Factor / quantity** | **Levels or definition** |
|---|---|
| Rack length $L_{SR}$ | 30, 60, 90 m |
| Motion profile M1 | $v_x$ = 2.5 m/s; $a_x$ = 1.5 m/s² |
| Motion profile M2 | $v_x$ = 4.0 m/s; $a_x$ = 3.0 m/s² |
| Visited picking locations n | 1–30 |
| Replications per scenario | 10 |
| Total scenarios | 180 |
| Total replication records | 1,800 |
| Principal comparison | 174 scenarios; n = 2–30 |
| Boundary set | 6 scenarios; n = 1 |
| Initial benchmark state | Fully occupied single-level rack; same benchmark initial-state definition |
| Maximum generated orders | 1,000 per replication (task-generator limit) |
| Warm-up | None; finite-horizon complete-cycle benchmark |
| Randomisation | Eligible locations sampled without replacement; separate random-number streams across replications |
| Accepted complete cycles | 5–133 per replication in the final data; median 11 |
| End-of-run handling | Incomplete end-cutoff cycle excluded from the replication mean |

### *4.2 Discrepancy and uncertainty measures*

Cycle-time discrepancy is reported in the simulation-minus-analytical direction, $\Delta CT = CT_S - CT_A$, in seconds and as a percentage of $CT_A$. Absolute relative discrepancy reports the size of the percentage difference without its sign. For the 174 principal scenarios, the summary includes mean, median and maximum absolute relative discrepancy. Mean signed error, mean absolute error (MAE) and root-mean-square error (RMSE) are also reported in seconds so that the comparison is not based on a percentage measure alone.

No numerical tolerance or equivalence margin was defined before the benchmark. The analysis therefore reports the size and direction of the differences and does not classify scenarios as passing or failing an equivalence test.

### *4.3 Derived throughput*

Analytical throughput is calculated from $CT_A$ and simulation throughput from $CT_S$ using the same one-item-per-visited-location relation. Consequently, for a fixed n, $CT_S < CT_A$ implies $TH_S > TH_A$. The throughput comparison is retained as an operational interpretation of the cycle-time result, not as an independent validation measure.

### *4.4 Robust factorial analysis*

The secondary factor analysis uses the 1,740 replication-level cycle-time results from the n > 1 scenarios. Rack length, motion profile and visited-location count are treated as categorical factors in a full-factorial linear model containing all two-way and three-way interactions. Sum-to-zero contrasts are used for factor coding. Inference uses HC3 heteroskedasticity-consistent covariance estimates and joint Wald chi-square tests. Holm adjustment is applied across the seven non-intercept tests. The statistical tests support the engineering trends; they are not used as a substitute for the discrepancy measures.

## 5. Results

### *5.1 Overall analytical–DES correspondence*

Across the 174 scenarios with n > 1, the mean absolute relative cycle-time discrepancy is 0.8655%, the median is 0.6758% and the maximum is 7.0370%. $CT_S$ is below $CT_A$ in all 174 scenarios. In seconds, the mean signed difference is −2.411 s, the MAE is 2.411 s and the RMSE is 2.744 s. The signed differences range from −6.601 to −0.598 s. The equal magnitudes of the mean signed difference and MAE result from the common negative direction of all principal discrepancies.

The main result is therefore close agreement in magnitude together with a consistent simulation-minus-analytical direction. The analytical and simulation values are not identical and the differences are not distributed around zero. The replication-based confidence intervals show that the simulation means themselves are estimated with sub-second average half-widths, so the observed directional discrepancy is not explained simply by imprecise scenario-mean estimation.

Replication-based uncertainty is small relative to the overall cycle-time range. Across the 174 principal scenarios, the 95% confidence-interval half-width for $CT_S$ has a mean of 0.590 s, a median of 0.534 s, and a maximum of 1.630 s. Representative scenario means and 95% intervals are: 30 m/M1/n = 2, 55.703 s [55.240, 56.167]; 60 m/M1/n = 8, 192.871 s [192.047, 193.695]; and 90 m/M2/n = 30, 585.385 s [584.802, 585.967]. The analytical CT lies inside the corresponding simulation-mean 95% interval in 3 of 174 scenarios. This count is descriptive only: the confidence intervals quantify uncertainty in the simulated mean and are not used as an equivalence margin or a pass/fail test of model correspondence.

Table 4. Cycle-time discrepancy by rack length and motion profile for n > 1.

| Rack length (m) | Motion | Scenarios | Mean \|ΔCT\| (%) | Median \|ΔCT\| (%) | Max \|ΔCT\| (%) | Mean ΔCT (s) |
|---|---|---|---|---|---|---|
| 30 | M1 | 29 | 0.959 | 0.887 | 3.627 | -2.817 |
| 30 | M2 | 29 | 0.811 | 0.750 | 2.705 | -2.245 |
| 60 | M1 | 29 | 0.912 | 0.680 | 5.283 | -2.634 |

| Rack length (m) | Motion | Scenarios | Mean \|ΔCT\| (%) | Median \|ΔCT\| (%) | Max \|ΔCT\| (%) | Mean ΔCT (s) |
|---|---|---|---|---|---|---|
| 60 | M2 | 29 | 0.842 | 0.670 | 4.163 | -2.411 |
| 90 | M1 | 29 | 0.885 | 0.553 | 7.037 | -2.317 |
| 90 | M2 | 29 | 0.784 | 0.540 | 5.126 | -2.044 |
| All | All | 174 | 0.8655 | 0.6758 | 7.0370 | -2.411 |

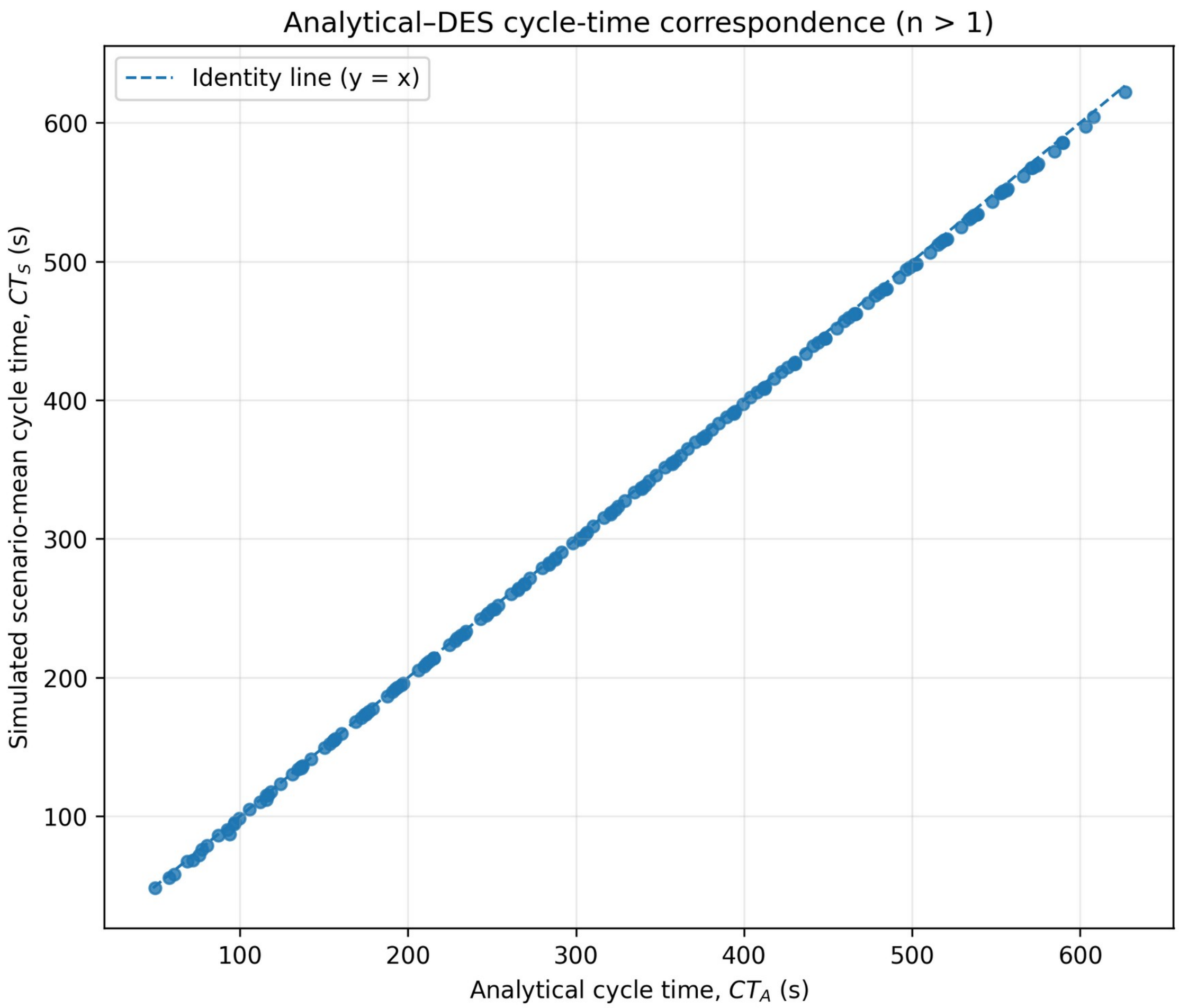

*Figure 2. Analytical cycle time versus simulated scenario-mean cycle time for the 174 principal scenarios. The dashed line is the identity line y = x.*

The parity plot places the 174 scenario means close to the identity line over the complete cycle-time range. A descriptive linear fit gives $R^2 = 0.999966$. This statistic is used only to describe the near-linear relation between the two sets of cycle times and is not an equivalence test.

### *5.2 Dependence on rack length, motion profile and n*

The mean absolute relative discrepancy remains below 1% for each of the six rack-length and motion-profile combinations. Across all six combinations, the absolute relative discrepancy at n = 30 is lower than at n = 2, although the values do not decrease monotonically at every intermediate n. Figure 3 shows the signed relative differences; all values remain below zero because $CT_S$ is lower than $CT_A$ throughout the principal set.

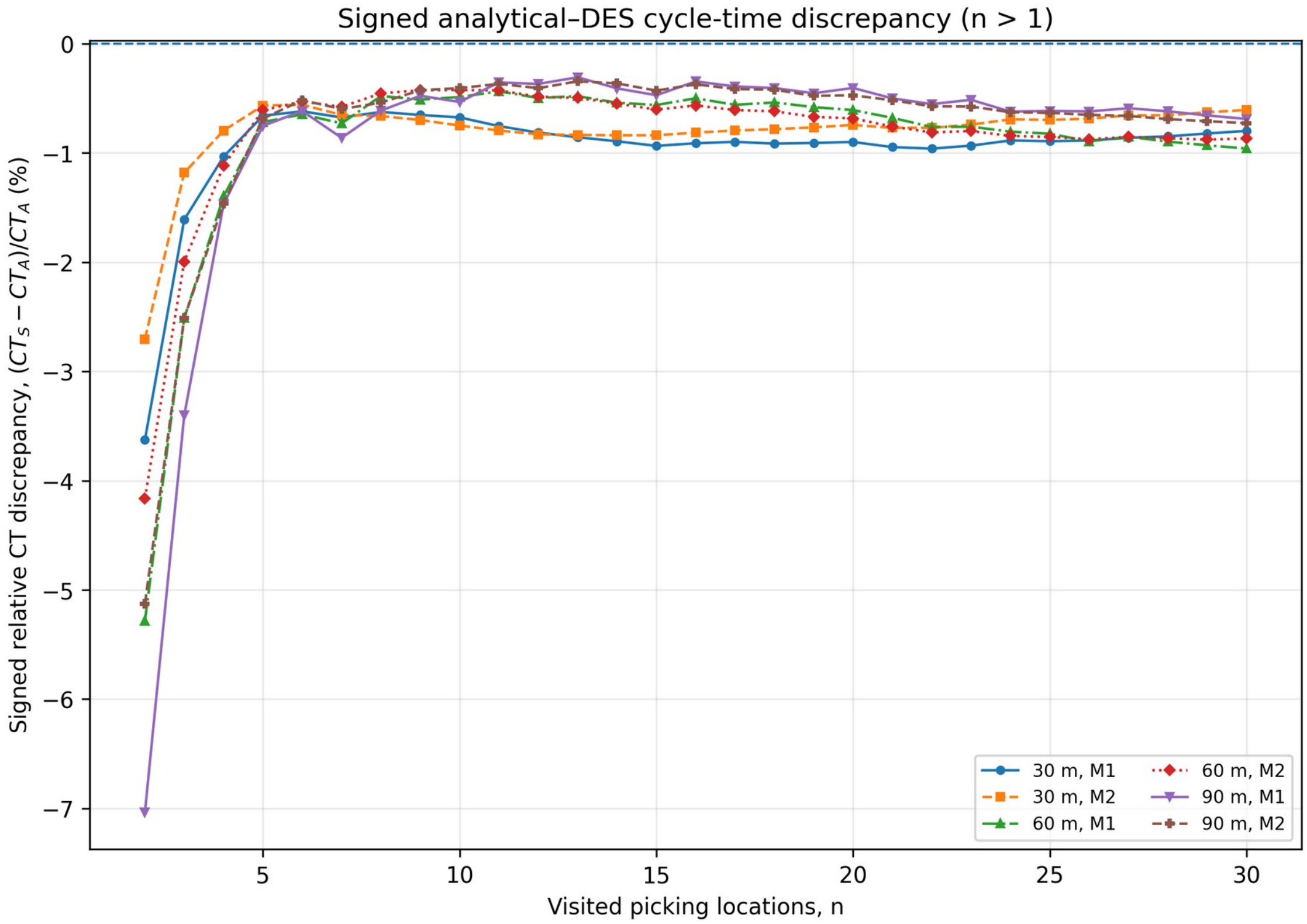

*Figure 3. Signed relative cycle-time discrepancy as a function of visited picking locations for each rack-length and motion-profile combination. Negative values indicate $CT_S < CT_A$.*

For the 30 m rack with the 2.5 m/s, 1.5 m/s² motion profile and n = 2, $CT_A$ = 57.800 s/cycle and $CT_S$ = 55.703 s/cycle, giving a signed difference of −2.097 s/cycle. The derived throughput difference is +4.689 items/hour, or +3.76% relative to $TH_A$. For the 90 m rack with the 4.0 m/s, 3.0 m/s² profile and n = 30, $CT_A$ = 589.677 s/cycle and $CT_S$ = 585.385 s/cycle, a difference of −4.292 s/cycle. The derived throughput difference is +1.343 items/hour, or +0.73%. These examples show why absolute and relative differences should be considered together.

### *5.3 Single-location boundary cases*

The six n = 1 scenarios are reported separately because their analytical cycle contains no travel-between term. For the 30 m rack with the 2.5 m/s, 1.5 m/s² profile, $CT_A$ = 37.333 s/cycle and $CT_S$ = 33.405 s/cycle, giving a difference of −3.928 s/cycle and a derived throughput difference of +11.76%. Across the six n = 1 cases, the mean absolute relative cycle-time discrepancy is 13.387% and the maximum is 18.480%. These cases are treated as structural boundary cases, not statistical outliers.

### *5.4 Performance trends and factor analysis*

The benchmark simulation results follow consistent engineering trends. Within every rack-length and motion-profile combination, cycle time increases as the number of visited locations increases. At fixed n and motion profile, longer racks produce higher cycle times. At fixed rack length and n, the faster motion profile produces lower cycle times. Derived item throughput increases with n over the tested range and, at fixed n, is higher for shorter racks and for the faster motion profile.

All seven terms in the HC3/Holm factor analysis remain statistically significant after adjustment (Table 5). The factor tests identify statistically detectable differences across rack length, motion profile, n and their interactions. The engineering interpretation nevertheless remains based on the size and direction of the cycle-time changes and the analytical–simulation discrepancy rather than on p-values alone.

Table 5. HC3-robust joint Wald tests for replication-level simulated cycle time ($n > 1$).

| Model term | Wald $\chi^2$ | df | Holm-adjusted p |
|---|---|---|---|
| Rack length | 525,079.215 | 2 | <0.001 |
| Motion profile | 245,269.688 | 1 | <0.001 |
| Visited picking locations | 42,925,388.120 | 28 | <0.001 |
| Rack × Motion | 24,101.393 | 2 | <0.001 |
| Rack × Visited locations | 5,518.951 | 56 | <0.001 |
| Motion × Visited locations | 6,397.459 | 28 | <0.001 |
| Rack × Motion × Visited locations | 103.035 | 56 | <0.001 |

## 6. Discussion and Limitations

### *6.1 Close correspondence with a systematic directional difference*

The analytical–DES benchmark shows close numerical correspondence between the analytical reference and the simulation model under aligned geometry, motion, handling and KPI definitions. The 0.8655% mean absolute relative discrepancy is accompanied by an MAE of 2.411 s and an RMSE of 2.744 s. The maximum relative discrepancy of 7.0370% shows that the small average difference does not mean that every scenario has the same level of agreement.

The sign of the difference is also consistent: $CT_S$ is lower than $CT_A$ in all 174 principal scenarios, and the derived $TH_S$ is correspondingly higher. The result is therefore better described as close correspondence with a systematic directional difference than as random variation around the analytical value. No formal equivalence margin was defined, so the comparison does not establish statistical equivalence.

### *6.2 Plausible origin of the residual difference*

A plausible contributor to the remaining difference is the way horizontal travel is represented. The analytical model evaluates the travel-time function at expected one-way and travel-between distances, whereas the simulation executes location-specific movements. The triangular branch of Equation (3) is concave in distance and the trapezoidal branch is linear with the same slope at the transition, so the complete travel-time function is concave. For a common travel-distance distribution, Jensen's inequality therefore gives $E[t(S)] \le t(E[S])$. This provides a theoretical mechanism by which evaluating travel time at an expected distance can be at least as large as averaging travel times over realised distances, which is consistent with the observed $CT_S < CT_A$ direction.

This argument does not prove that Jensen's inequality alone causes the 174/174 directional result. The simulation samples eligible locations without replacement and then applies the side/column controller rule, so its realised movement distribution is not guaranteed to be identical to the distance distribution implicit in the analytical derivation. Other implementation details may also contribute. A follow-up controlled experiment that holds the sampled locations fixed while changing only the distance or sequencing representation would be required to isolate the causal contribution of each mechanism.

### *6.3 Why relative discrepancy is larger for small n*

The larger relative discrepancies at small n are consistent with the cycle-time structure. Every visited location adds 16.8 s of tote handling and item picking. As n increases, this repeated contribution forms a larger part of the complete cycle, so a travel-related difference of a few seconds represents a smaller percentage of the total. This is consistent with the lower endpoint discrepancy at n = 30 than at n = 2 in all six rack/motion combinations, although the intermediate values are not strictly monotonic.

The n = 1 scenarios also have a different cycle structure because the travel-between contribution is zero. The one-way travel representation therefore accounts for a larger share of their complete cycle, which is why they are reported separately from the multi-location scenarios.

### *6.4 Implications for analytical and simulation modelling*

Within the tested single-level conditions, the analytical formulation provides a compact way to estimate cycle time and derived item throughput, while DES retains the detail needed for explicit location-specific movement and controller logic. The benchmark therefore contributes a reproducible way to examine model-form correspondence, not merely a check that two implementations return similar averages. In particular, it shows that a mean absolute relative discrepancy below 1% can coexist with a persistent directional difference and with larger local deviations at short cycles. This distinction matters when analytical approximations are used as performance references for more detailed simulation models.

This role is consistent with analytical–simulation studies in other automated-storage architectures. Eder (2024) and Trost and Eder (2024) compare analytical performance formulations with DES for shuttle-based and robotic compact storage/retrieval systems. The present study applies the same general type of comparison to the intra-aisle integrated-picking cycle and additionally documents the distance interpretation needed to reproduce the analytical reference values.

The comparison does not show that the simulation model is more correct than the analytical model. Neither model was adjusted to match an operating warehouse. A real-system validation would require representative warehouse data and comparison with measured travel, handling, picking or order-performance results (Sargent, 2013; Ferrari et al., 2024).

### *6.5 Reproducibility*

The reproducibility contribution has two parts. On the analytical side, the reconstruction states $d_{ES} = 1.5$ m and $d_{TB} = 0$ m and reproduces the published cycle-time values and rounded throughput values. On the simulation side, the retained benchmark files contain the 180 scenario results and the 1,800 replication results from which the scenario means, uncertainty measures and discrepancy statistics can be recalculated. The HC3/Holm analysis can likewise be reproduced from the 1,740 replication-level results for $n > 1$ using the stated categorical full-factorial specification and sum-to-zero contrasts.

This separation of analytical assumptions, simulation implementation, experiment design and exported results is also consistent with the reporting principles of the STRESS guidelines (Monks et al., 2019).

### *6.6 Limitations*

The conclusions apply only to the tested rack lengths, motion profiles, n values and fixed handling/picking times. No equivalence margin was defined, and the results therefore do not establish formal statistical equivalence. The analytical and simulation representations are aligned but not identical, and the present experiment does not isolate the cause of the residual directional difference. Both throughput values are derived from cycle time and therefore do not provide an independent validation signal. The model was not calibrated against an operating warehouse, so predictive validity for a real facility is outside the scope of this study. The six $n = 1$ cases are reported separately because their cycle structure differs from the $n > 1$ cases.

## 7. Conclusion

This article reports a single-level analytical-reference comparison for intra-aisle robotic picking. The analytical multi-command cycle of Lerher et al. (2025) is reconstructed and compared with the corresponding benchmark simulation results using the same rack geometry, motion and handling parameters and the same cycle-time and throughput definitions.

Reproducing the published analytical tables requires $d_{ES}$ = 1.5 m and $d_{TB}$ = 0 m. The analytical–DES benchmark comprises 180 scenarios with ten simulation replications per scenario. The six n = 1 cases are reported separately, leaving 174 multi-location scenarios for the principal comparison.

For those 174 scenarios, the mean absolute relative cycle-time discrepancy is 0.8655%, the median is 0.6758% and the maximum is 7.0370%. The MAE is 2.411 s and the RMSE is 2.744 s. $CT_S$ is below $CT_A$ in all 174 scenarios. The results therefore show small average differences but not exact agreement. The systematic direction is consistent with a residual difference between the analytical distance representation and the location-specific simulated path, although the present study does not isolate one causal mechanism.

Within the tested assumptions, the analytical formulation is a compact reference for cycle-time and derived-throughput calculations, while DES retains the detail needed for explicit movement and control logic. The comparison establishes analytical–simulation correspondence, not validation against a physical warehouse. Future work should separate the sources of the remaining directional difference, extend the tested parameter range and compare model results with measured warehouse data.

## Declarations

Authorship. Mohsen Reda Mahmoud Moursi and Ahmed Stohy are the authors of this manuscript.

Disclosure statement. The authors report no potential conflict of interest.

**Funding.** This research received no specific grant from any funding agency in the public, commercial, or not-for-profit sectors.

Data availability. The scenario-level and replication-level research data supporting the findings of this study, together with the analysis code used to reproduce the reported descriptive and robust factorial analyses, are available from the corresponding author upon reasonable request. The Siemens Tecnomatix Plant Simulation model is also available from the corresponding author upon reasonable request, subject to software-access requirements.

Thesis origin. This article is derived from a focused part of the first author’s 2026 Master’s thesis, “Modelling and Simulation of Intra-Aisle Picking Operations in Automated Warehousing” (Moursi, 2026).

## References

Boysen, N., de Koster, R., & Weidinger, F. (2019). Warehousing in the e-commerce era: A survey. European Journal of Operational Research, 277(2), 396–411. https://doi.org/10.1016/j.ejor.2018.08.023

de Koster, R., Le-Duc, T., & Roodbergen, K. J. (2007). Design and control of warehouse order picking: A literature review. European Journal of Operational Research, 182(2), 481–501. https://doi.org/10.1016/j.ejor.2006.07.009

Eder, M. (2020). Analytical model to estimate the performance of shuttle-based storage and retrieval systems with class-based storage policy. The International Journal of Advanced Manufacturing Technology, 107, 2091–2106. https://doi.org/10.1007/s00170-020-04990-y

Eder, M. (2024). Analytical examination of shuttle-based storage and retrieval systems with multiple-capacity lifts. The International Journal of Advanced Manufacturing Technology, 133, 5053–5064. https://doi.org/10.1007/s00170-024-14058-w

Ferrari, A., Carlin, A., Rafele, C., & Zenezini, G. (2024). A method for developing and validating simulation models for automated storage and retrieval system digital twins. The International Journal of Advanced Manufacturing Technology, 131, 5369–5382. https://doi.org/10.1007/s00170-023-12660-y

Ferrari, A., Campanale, C., Carlin, A., & Mangano, G. (2025). Performance assessment of a multi-level shuttle system: An optimisation-embedded simulation study on time and energy consumption. Journal of Simulation, 19(4), 411–426. https://doi.org/10.1080/17477778.2024.2404215

Law, A. M. (2020). Statistical analysis of simulation output data: The practical state of the art. In Proceedings of the 2020 Winter Simulation Conference (pp. 1117–1127). IEEE. https://doi.org/10.1109/WSC48552.2020.9383993

Lerher, T., Marolt, J., Bizjak, L., & Jerman, B. (2025). Autonomous mobile robots AMRs equipped with picking robot for order-picking in storage rack. In Proceedings of the 17th International Material Handling Research Colloquium (IMHRC), Paper 16, Trondheim, Norway.

Monks, T., Currie, C. S. M., Onggo, B. S., Robinson, S., Kunc, M., & Taylor, S. J. E. (2019). Strengthening the reporting of empirical simulation studies: Introducing the STRESS guidelines. Journal of Simulation, 13(1), 55–67. https://doi.org/10.1080/17477778.2018.1442155

Moursi, M. R. M. (2026). Modelling and Simulation of Intra-Aisle Picking Operations in Automated Warehousing [Master's thesis, Graz University of Technology].

Saini, R., Vaidya, O. S., & Venkitasubramony, R. (2025). Simulation to analyse warehouse operational performance: A systematic literature review. Journal of Simulation, 19(4), 455–481. https://doi.org/10.1080/17477778.2025.2485231

Sargent, R. G. (2013). Verification and validation of simulation models. Journal of Simulation, 7(1), 12–24. https://doi.org/10.1057/jos.2012.20

Tappia, E., Roy, D., Melacini, M., & de Koster, R. (2019). Integrated storage-order picking systems: Technology, performance models, and design insights. European Journal of Operational Research, 274(3), 947–965. https://doi.org/10.1016/j.ejor.2018.10.048

Trost, P., & Eder, M. (2024). An analytical approach for the performance calculation of an RCS/RS with several picking stations. The International Journal of Advanced Manufacturing Technology, 134, 627–644. https://doi.org/10.1007/s00170-024-14124-3